# ONOT/TONO-Print: A Paired Benchmark for Identity-Document Portrait Quality under Document Production Shift

*Alexander Vinogradov*
*IU International University of Applied Sciences*
*Dataset:*
*https://github.com/tehnix53/onot-tono-printed*

**Abstract** - Identity-document portraits are modified during document production by printing, engraving, and integration into the document template. These transformations may change image texture, contrast, colour, and other visual properties without necessarily changing the underlying biometric or presentation information.
In this work, we introduce ONOT/TONO-Print, a small paired benchmark for evaluating face-image quality models under identity-document production shift. The benchmark is based on ONOT and TONO portraits and contains synthetic variants representing five common production techniques: toner, inkjet, dye sublimation, laser engraving, and thermal transfer. A smaller physical subset contains portraits printed and scanned using inkjet and toner devices.
The benchmark is used to compare the behaviour of biometric and ICAO-related face-quality components before and after production. Results show that production can introduce substantial domain shift for some quality estimators, while other models remain comparatively stable.

## 1. Introduction

Face image quality models are commonly evaluated on digital face images. In identity documents, however, the portrait is not used only in its original digital form. It may be printed, engraved, or otherwise transferred to the physical document and combined with document-specific security and design elements.

These production processes can change local texture, contrast, colour, and sharpness. As a result, a quality model may respond to characteristics introduced by the document production process rather than to changes in the underlying face information.

ONOT (Di Domenico et al., 2024) and TONO (Borghi et al., 2025) provide controlled portrait data for studying ICAO-related face presentation characteristics. ONOT contains compliant portraits, while TONO contains corresponding examples with controlled presentation violations such as head coverings, sunglasses, non-neutral expression, non-frontal pose, gaze deviation, and closed eyes.

In this work, we extend this setting with controlled production transformations and a small physical print-and-scan subset. The goal is not to reproduce every identity-document manufacturing process exactly, but to provide a paired benchmark for analysing the robustness of face-quality measures under production-related domain shift.

## 2. Dataset description

The benchmark is based on 256 identities from ONOT and corresponding TONO portraits. For each selected portrait, synthetic production variants were generated using five transformations:

- toner;
- inkjet;
- dye sublimation;
- laser engraving;
- thermal transfer.

The transformations approximate characteristic visual effects of the corresponding production techniques while preserving the identity and presentation condition of the source portrait (Figure 1).

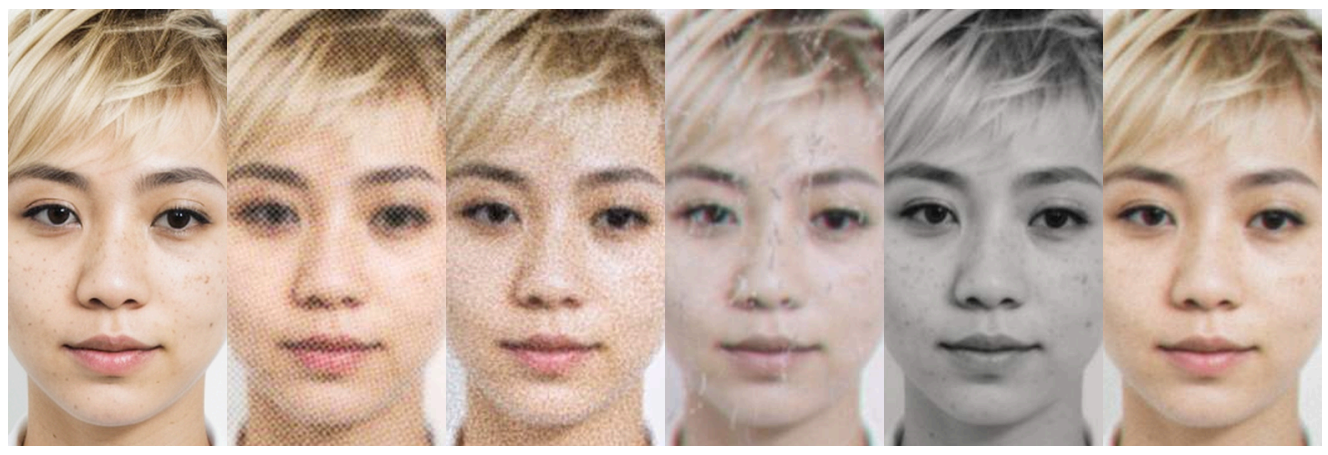

a b c d e f

Fig. 1 Example synthetic production variants of the same source portrait: (a) digital source, (b) toner, (c) inkjet, (d) dye sublimation, (e) laser engraving, and (f) thermal transfer.

A smaller physical subset was additionally created using real inkjet and toner printing followed by scanning (Figure 2). This subset is used as a physical sanity check and is not intended to represent the full range of identity-document manufacturing technologies.

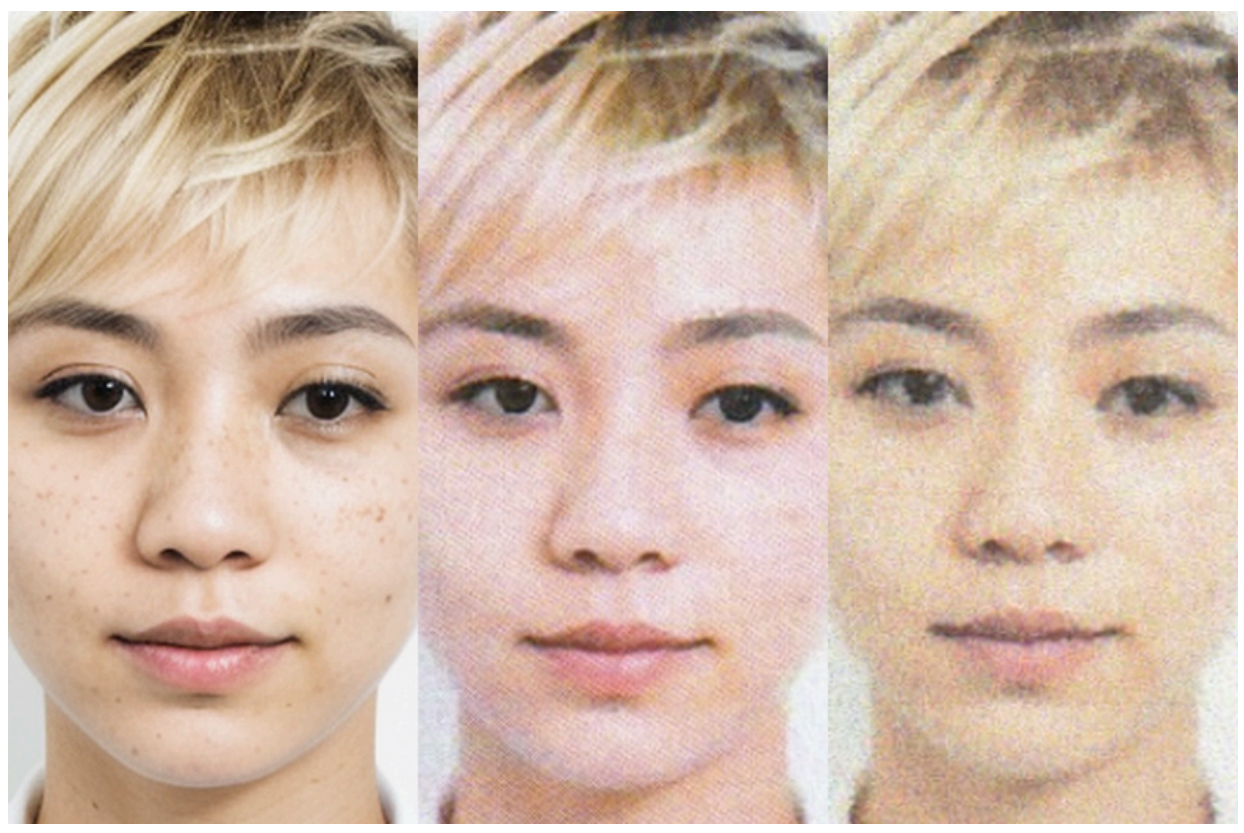

a b c

Fig. 2 Example physical print-and-scan variants of the same source portrait: (a) digital source, (b) toner, and (c) inkjet.

Each transformed sample is linked to its corresponding digital source using the original identity naming from ONOT dataset. ONOT/TONO-Print dataset structure is presented in Table 1.

Table 1 - ONOT/TONO-Print dataset structure

| Split | Content | N |
|---|---|---|
| synthetic / ONOT | bona fide × 5 techniques | 256 × 5 |
| synthetic / TONO | ICAO-related violations × 5 techniques | 1707 × 5 |
| physical / ONOT | bona fide × 2 techniques | 256 × 2 |
| physical / TONO / toner | ICAO-related violations | 1707 |
| physical / TONO / inkjet | head covering + sunglasses only | 281 + 50 |

## 3. Experiments

The benchmark was used to measure production-related domain shift in face-image quality models. We evaluate OFIQ UQS (Merkle et al., 2024), CR-FIQA (Boutros et al., 2023), MagFace (Meng et al., 2021), eDifFIQA (Babnik et al., 2024), and FaceQAN (Babnik et al., 2022).

For each method, the digital portrait is used as the reference. Production sensitivity is measured as the change between the digital and transformed condition. The results show that the effect of production is strongly model-dependent (Table 2).

Table 2 - Relative score change of face-quality models after portrait production

| Technique | ΔQ = 100 × (Qprinted − Qsource) / Qsource | | | | |
|---|---|---|---|---|---|
| | OFIQ | eDif FIQA | CR-FIQA | Mag Face | Face QAN |
| ONOT digital vs synthetic production | | | | | |
| dye | -30.2 | -7.0 | -2.7 | +1.7 | -15.8 |
| inkjet | -30.4 | -4.0 | -3.9 | -3.0 | +3.7 |
| laser engraved | -31.2 | -5.6 | -2.9 | -6.1 | -6.6 |
| thermal transfer | -6.1 | -2.4 | -0.7 | +1.4 | -2.1 |
| toner | -48.5 | -5.5 | -6.9 | -4.6 | -2.9 |
| ONOT digital vs physical print-scan | | | | | |
| inkjet | -53.3 | -5.2 | -6.4 | +9.5 | -8.4 |
| toner | -39.2 | -4.9 | -4.4 | +6.3 | -2.4 |

## 4. Conclusions and Future Work

ONOT/TONO-Print provides a small paired benchmark for studying face-image quality under identity-document production shift. The benchmark separates the original portrait content from production-related transformations and allows direct comparison of quality scores and component-level performance before and after production.
The results show that production-related degradation is not reflected consistently across face-quality models. Some methods remain relatively stable, while others show substantial score changes under the same transformations.
The current benchmark is limited to five simulated production techniques and a small physical inkjet/toner print-and-scan subset. It does not reproduce the full physical complexity of document manufacturing, substrates, laminates, or security elements. Future work may extend the benchmark with additional physical production techniques and document types.